\documentclass[letterpaper, 10 pt, conference]{ieeeconf}  

\IEEEoverridecommandlockouts                              

\usepackage{graphicx,subcaption} 

\usepackage{algorithm, amsmath,amssymb}
\usepackage{algpseudocode}
\usepackage{booktabs}

\makeatletter               
\let\NAT@parse\undefined
\makeatother
\usepackage{hyperref}
\hypersetup{colorlinks,breaklinks,
linkcolor=[rgb]{0.5,0.,0.},
citecolor=[rgb]{0.000,0.427,0.173},
urlcolor=[rgb]{0.031,0.318,0.612}}

\title{\LARGE \bf
Designing Tools with Control Confidence*
}

\author{%
  Ajith Anil Meera$^{1}$,%
  \quad Abian Torres$^{2, 3}$,%
  \quad Pablo Lanillos$^{2,1}$
  \thanks{*This research is supported by the Metatool project (Grant agreement 101070940) under the EIC pathfinder program. Corresponding author: {\tt\small ajith.anilmeera@donders.ru.nl}}%
  \thanks{$^{1}$Donders Institute, Radboud University Nijmegen, The Netherlands}%
  \thanks{$^{2}$Cajal Neuroscience Center, Spanish Research Council}%
  \thanks{$^{3}$Polytechnic University of Madrid, Madrid, Spain}%
}

\begin{document}

\maketitle
\thispagestyle{empty}
\pagestyle{empty}

\begin{abstract}

Prehistoric humans invented stone tools for specialized tasks by not just maximizing the tool's immediate goal-completion accuracy, but also increasing their confidence in the tool for later use under similar settings. This factor contributed to the increased robustness of the tool, i.e., the least performance deviations under environmental uncertainties. However, the current autonomous tool design frameworks solely rely on performance optimization, without considering the agent's confidence in tool use for repeated use. Here, we take a step towards filling this gap by i) defining an optimization framework for task-conditioned autonomous hand tool design for robots, where ii) we introduce a neuro-inspired control confidence term into the optimization routine that helps the agent to design tools with higher robustness. Through rigorous simulations using a robotic arm, we show that tools designed with control confidence as the objective function are more robust to environmental uncertainties during tool use than a pure accuracy-driven objective. We further show that adding control confidence to the objective function for tool design provides a balance between the robustness and goal accuracy of the designed tools under control perturbations. Finally, we show that our CMAES-based evolutionary optimization strategy for autonomous tool design outperforms other state-of-the-art optimizers by designing the optimal tool within the fewest iterations. Code: \url{https://github.com/ajitham123/Tool_design_control_confidence}.

\end{abstract}

\section{INTRODUCTION}
Autonomous tool design is a very promising field of research for robotic and industrial applications \cite{qin2023robot}. Exploiting the full power of 3D printers and computer-aided design, new solutions are progressively automating the process, but a human-in-the-loop is still a necessity. The major challenge for autonomous design in robotics is to ensure that the proposed tool is actually useful and reliable for the task. Current automatic solutions focus on performance optimization (i.e., maximum accuracy or reward) and often disregard the tool's robustness and reliability in the presence of perturbations (e.g., unmodeled changes in the environment dynamics). Learning methods capture these environmental uncertainties using a data-intensive approach that attempts to encode all possible alterations \cite{li_learning_2023,liu_learning_2023,yuan_transform2act_2022}. However, acquiring complete knowledge of world uncertainties is usually intractable. On the other hand, humans are very good at tool use, design, and invention, making it a hallmark of intelligent behavior~\cite{amant2005tool}, indicating the great potential of neuro-inspired tool design. 

\begin{figure}[!t]
\centering
\includegraphics[scale = 0.45]{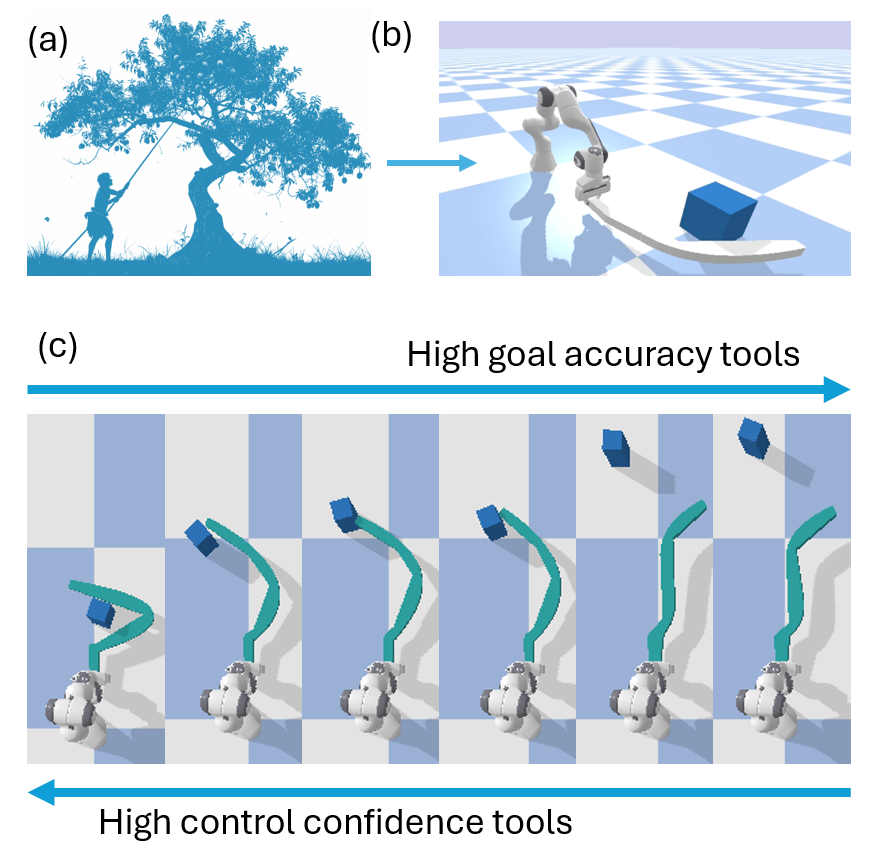}
\caption{Designing tools for high controllability: a) Depicts a prehistoric human performing a manipulation task with a stick tool, b) Our PyBullet simulation environment mimicking (a) with a manipulator arm and a tool to manipulate a box on the ground, c) The final results of our tool design algorithm for six different weights on control confidence and box goal accuracy. High control confidence tools are curved to maximize the controllability of the box at the expense of accuracy of the box goal position.  }
\label{fig:teaser}
\end{figure}

Taking inspiration from recent developments in computational neuroscience in modeling brain metacognition \cite{fleming2024metacognition}, here we propose an alternative approach to tool design (Fig. \ref{fig:teaser}), which includes, in addition to accuracy maximization, maximization of confidence that the robot has to control the tool. Mathematically, control confidence can be expressed as the posterior precision (inverse of the covariance matrix) of the agent's control signal during tool use~\cite{meera2024confidence}, which describes the tool's controllability. Maximizing it endows the agent with the ability to deal with perturbations in the control signal during the use of the designed tool, thereby improving the robustness of the tool. However, confidence alone may not provide the tool design that maximizes goal accuracy for a particular task. Thus, we use a mathematical objective function that seemingly balances robustness and goal accuracy. To derive this objective, we rely on the mathematical foundations of the Free Energy Principle (FEP)~\cite{friston2010free, lanillos2021active} from neuroscience, where the brain's decision-making (tool design in our context) is entirely driven by minimizing the free energy objective.


\begin{figure*}[!hbtp]
\centering
\includegraphics[scale = 0.37]{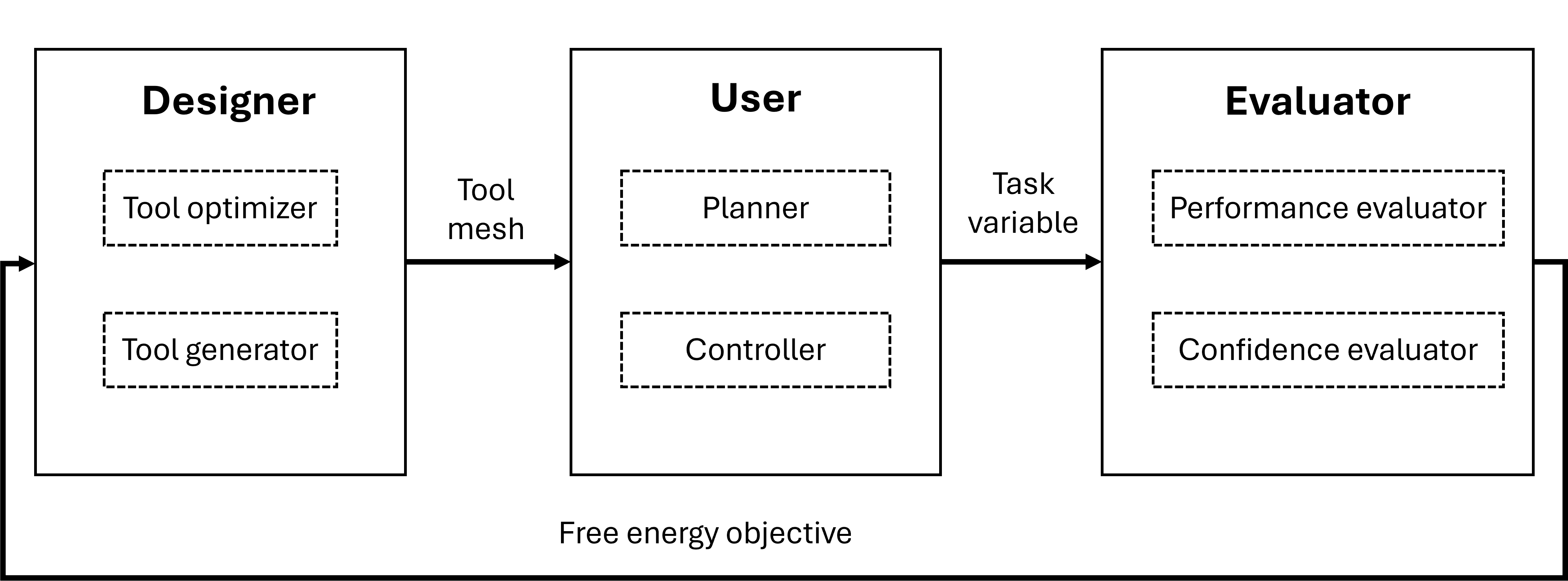}
\caption{The schematic of the tool design pipeline. The designer designs and optimizes a tool mesh design. This mesh is used by the user to perform a task in a simulation environment using a fixed plan and a standard robot controller. The task variable in the simulation is tracked by the evaluator and the tool performance is evaluated. A confidence value is attributed to each tool and is used to compute the free energy objective that the designer uses to optimize the tool design.}
\label{fig:schematic}
\end{figure*}




Within the scope of this work, we define the design of autonomous hand tool (as depicted in Fig.~\ref{fig:schematic}) as an optimization process~\cite{liu_learning_2023}, where i) a designer optimizes a parametric representation of a tool mesh, ii) a user agent uses this tool in the environment, and iii) an evaluator measures the tool's task performance and confidence. With this framework, we investigate whether tool design via the optimization of control confidence improves the tool's performance robustness against environmental uncertainties. We assume a known prior knowledge about the tool use. In the same way that for designing a spoon for coffee mixing, we know the swirling motion of the tool, we fix the planner for the robot manipulation task (object pulling and pushing task). This turns our research focus to the evaluator and the designer's optimizer.


Through rigorous simulations (in PyBullet) with a 7DOF robotic manipulator (Franka), we show that $i)$ the tools designed by control confidence as the objective function present the highest reliability under environmental perturbations (e.g., object mass change); $ii)$ using the free energy objective (which includes control confidence) balances goal accuracy and robustness; and iii) we statistically show that our CMAES-based evolutionary optimization outperforms other state-of-the-art optimizers in optimizing tool design the fastest.


\section{RELATED WORK}
Prior work on autonomous tool design can be categorized based on (i) how tools are represented and generated, (ii) which objective functions are used, (iii) which optimizers are used, (iv) how control is learned, and (v) how task conditioning is performed alongside task constraints (e.g., attachment constraints). Importantly, the tool design literature has focused on generating plausible designs without a task~\cite{wu2021deepcaddeepgenerativenetwork} or optimizing parametric designs to maximize task success. Here, we focus on the latter, emphasizing that the literature does not explicitly account for the reliability of the tool in the presence of unmodeled uncertainties or perturbations.

\subsection{Representations and generation}
Depending on how the parts of the tool are defined, we can find constructive approaches for assembling tools from available parts and attachments, and synthesis methods for creating novel tool morphologies. 

\subsubsection{Tool construction}
In symbolic and graph formulations, parts and relations are modeled as discrete structures and assemblies are composed by checking a small set of geometric relations and compatibilities \cite{wicaksono_towards_2017,wicaksono_cognitive_2020,yang_autonomous_2020}. Related approaches assemble parts using task-driven heuristics and simple contact affordances \cite{choi_creating_nodate,yang_autonomous_2020}. The emphasis is on feasible assemblies rather than novel morphologies. Tools can also be constructed from available parts either by geometric shape matching and feasibility checks \cite{nair_tool_2019} or by feature-guided heuristic search that exploits object properties during planning \cite{nair_feature_2020}.
 
\subsubsection{Tool synthesis}
Tool shapes are parameterized continuously and optimized using simulators \cite{li_learning_2023} or learned surrogates \cite{liu_learning_2023}. Latent generative models can be used to enable backpropagation of task-performance signals to guide morphology \cite{wu_imagine_2020,li_generating_2024,liu_learning_2023}.
Recently, large models allow programmatic computational designs or part assemblies that are then verified in simulation or with rapid physical tests \cite{lin_robotsmith_2025,gao_vlmgineer_2025,pun_generating_2025}. Our setup uses a compact parametric representation that facilitates optimization and fabrication, avoiding dependence on a specific synthesis language, but is limited by the representational power of the mathematical description.

\subsection{Objective formulation}
Success-only objectives (e.g., binary classification) or performance objectives (reward-based) optimize designs based on task success or closeness to success. They are common in tool design and tool-use pipelines \cite{wu_imagine_2020,lin_robotsmith_2025, liu_learning_2023} and in policy evaluation \cite{seita_toolflownet_2023}. However, optimizing pure task success or pure performance reward often yields solutions that are highly sensitive to unmodeled perturbations (e.g., slight changes in mass, friction, initial pose), resulting in a suboptimal solution during practical tool use.
From an objective-formulation perspective, active inference  \cite{collis_understanding_2024} supports composite objectives that balance utility (task success) with epistemic value (information gain). This formulation has been used to analyze tool discovery versus innovation. This motivates us to go beyond success-only criteria-- by coupling the performance term with a confidence/controllability term. Importantly, there is a literature gap for explicit mechanisms that can handle unmodeled environmental perturbations or uncertainties within their objective. Hence, we fill this gap by proposing a more comprehensive objective function for the designer that balances both task performance and the controllability of the tool under environmental perturbations.

\subsection{Design optimizers}
Tool design optimizers can be classified as gradient-based and gradient-free methods. Gradient-based optimization uses differentiable physics or surrogates to efficiently shape designs. However, they struggle with discontinuities caused by impacts and stick–slip contacts \cite{wu_imagine_2020,li_learning_2023}. 
Gradient-free search strategies (e.g. evolutionary methods) and policy-guided proposals \cite{xu_dynamics-guided_2025, yuan_transform2act_2022} avoid brittle gradients and operate naturally with non-smooth dynamics.
Our objective function for tool design is highly nonlinear, noisy, and nonconvex. Therefore, we propose the use of the state-of-the-art gradient-free optimizer Covariance Matrix Adaptation Evolution Strategy (CMA-ES) for our tool design problem.


\subsection{Tool use}


{\sloppy
The most common approach to tool use is based on classical control or learned policies for fixed tool geometries \cite{lu_dynamic_2025, qi_learning_2024, seita_toolflownet_2023}. Another approach is co-design, which integrates design and control, but is computationally expensive because controllers are re-trained per iteration \cite{li_roman_2022}. When differentiable simulators or surrogates are available, joint gradients are used to update the tool geometry and policy simultaneously \cite{wu_imagine_2020, li_learning_2023}. Since tool use isn't the primary focus of our work, we simplify it by using a standardized classical controller to follow a fixed robot motion plan. 
\par}

\subsection{Task Conditioning}
Task-conditioning-- generating tool design for a specific task and task constraints-- is one of the major challenges in autonomous tool designs. Previous approaches have used conditioning signals, such as prompts, affordance cues, demonstration trajectories, or goal states to bias the design to be task-relevant \cite{wu_imagine_2020,gao_vlmgineer_2025,lin_robotsmith_2025}. Assembly constraints and end-effector attachment are usually modeled explicitly so that proposed designs can be built and mounted \cite{kodnongbua_computational_2022,li_roman_2022}. Our approach uses task variables in the objective function to tie the tool design to the task. 
Besides, our design variables are geometric and do not take into account the material composition or microstructure.

In summary, this work combines compact parametric designs \cite{wu_imagine_2020}, a gradient-free search that remains robust under contact-rich dynamics \cite{xu_dynamics-guided_2025}, and explicit treatment of tool controllability/confidence.

\section{MATHEMATICAL FRAMEWORK}
\subsection{Problem definition}
We address the problem of designing a hand tool $\mathcal{T}$ for a robot manipulator arm to perform a task $\Gamma$. The tool is parametrized as a polynomial curve that is fully defined by $C = [c_1,\ldots, c_n, 0]$. The last term is set to 0 to ensure that the tool has a fixed starting point [0,0]. A straight segment is always added to the end of the tool for grasping (see Figure \ref{fig:teaser}). Inspired by the invention of hook-shaped tools by bending sticks, we restrict the tool shape to have a fixed arc length $L(C) = L_0$. For practicality, we constrained the task $\Gamma$ to reach, move, and pull a box (as shown in Figure \ref{fig:teaser}b) to a desired goal location, under a given robot motion plan $\pi_{plan}$. The robot follows a known motion primitive $\pi_{\mathrm{plan}}=\pi_0$ by holding the tool to generate the joint torques $U(\pi_{\mathrm{plan}})$. This is expected to drive the task variable $\tau(C,U)$ to the desired goal $\tau^g$ for successful completion of the task. Since the task is to move the box to the goal location $X^g$, $\tau(C,U) = X(t)$, where $X$ is the box position at time $t$. The design problem is framed as an optimization problem with an objective function $F$ as:
\begin{equation} \label{eqn:opti_general_form}
\mathcal{T}^* =  \underset{C \in \mathbb{R}^n}{\mathrm{optimize}} \quad \sum_{t=0}^T
F\big(X(t), X^g,t \big),    
\end{equation}
\[
\text{subject to} \quad 
L(C) = L_0, \ 
C = [c_1,\ldots, c_n, 0], \ \pi_{plan} = \pi_0.
\]
In the literature, $F$ is usually a reward or success function that describes a pure performance measure for the tool. We propose a brain-inspired objective function for tool design, which will be defined in the next section.

\subsection{Control Confidence} \label{sec:control_conf}
We describe the mathematical foundations of confidence modeling based on the principles of FEP and metacognition. Within the metacognition literature, there is a growing consensus on using the second-order statistic of the decision variable (confidence in the decision) as a measure of metacognition \cite{fleming2017self}. Recent developments in FEP has contributed to a mathematical description of the second-order statistic of control signal \cite{meera2024confidence,friston2008variational}. We combine both of these ideas within the scope of tool design by incorporating metacognition (through control confidence) into the objective function for tool design given in Equation \ref{eqn:opti_general_form}. 

We define control confidence as a measure of the second-order statistic of the control signal. Intuitively, it reflects the agent's confidence in its controller towards task completion under a given tool. According to active inference, the precision (posterior inverse covariance matrix) of the control signal is the second gradient of the internal energy $E$ of the agent. For our problem, $E$ is defined by:
\begin{equation} \label{eqn:internal_energy}
E(t) = \frac{1}{2}(X(t) - X^g)^\top P^X (X(t) - X^g),
\end{equation}
where $P^X \in \mathbb{R}^{3 \times 3}$ is a constant matrix. $E$ can be seen as a precision weighted squared goal error. The control confidence $\Pi^U$ is defined as the double gradient of $E$ with respect to control $U$ as \cite{meera2024confidence,friston2008variational}:
\begin{equation}
    \Pi^U = \frac{\partial^2 E}{\partial U^2},
\end{equation}
which can be evaluated by differentiating $E$ twice:
\begin{equation}
\frac{\partial E}{\partial U}
= (X - X^g)^\top P^X \frac{\partial X}{\partial U},
\end{equation}
\begin{equation} \label{eqn:control_conf}
\Pi^U = \frac{\partial^2 E}{\partial U^2}
= \frac{\partial X}{\partial U}^\top P^X \frac{\partial X}{\partial U}
+ (X - X^g)^\top P^X \frac{\partial^2 X}{\partial U^2}.
\end{equation}
The first term in $\Pi^U$ acts as a task-directed empowerment measure (or controllability), quantifying the influence of control inputs on the task-relevant state directions emphasized by $P^X$. The second term rewards high control sensitivity away from the goal (strong reactions) and low control sensitivity (small corrections) around the goal, promoting stability around the goal. It also acts as an exploratory term that promotes higher controllability even at the expense of not reaching the goal. This is particularly relevant for tool design because it contributes to the tool's adaptability/robustness while in use under uncertainties. In combination, the agent has a higher control confidence $\Pi^U$ in a tool in regions with greater controllability and in regions away from the goal with greater control sensitivity.


\begin{figure*}[!hbtp]
\centering
\includegraphics[scale = 0.5]{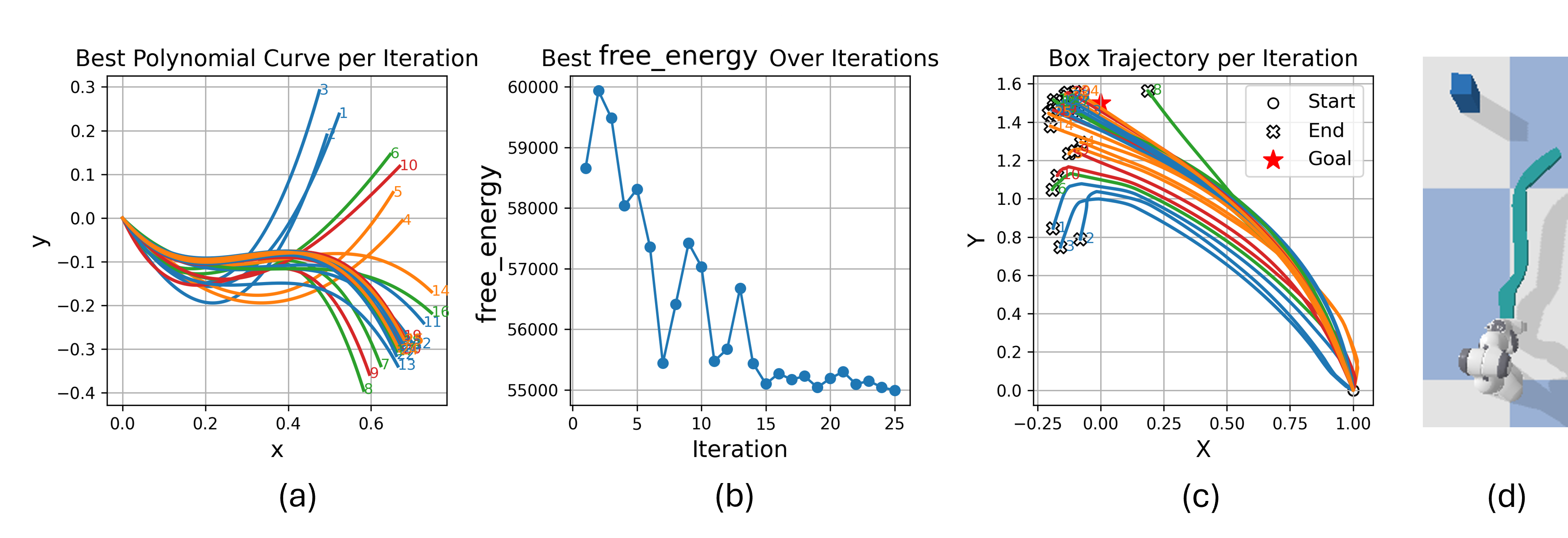}
\caption{ The realization of the optimization process of the tool design. a) The evolution of the best polynomial curves of the tool per iteration. b) The free energy for the best candidate in the population per iteration of the cmaes objective. c) The top view of the box movement trajectories from its start location (black circle) to the end location (black cross) while using the best tool per iteration. d) The final optimal tool design for $p^X=40$. }
\label{fig:cmaes_progress}
\end{figure*}

\subsection{Tool Optimization}
According to FEP, brain minimizes a quantity called free energy for decision making and other cognitive functions \cite{friston2010free}. Inspired by this idea, we formulate the tool design problem as the minimization of free energy objective $F$ written as:
\begin{equation} \label{eqn:free_energy}
    F = \frac{1}{2}(X - X^g)^\top P^X (X - X^g) - \frac{1}{2} \log |\Pi^U|.
\end{equation}
$F$ is composed of two terms: i) the weighted goal error $E$ from Equation \ref{eqn:internal_energy} that indicates the performance of the tool for the task, and ii) the control confidence on the tool (information entropy of $\Pi^U$). Minimizing free energy minimizes goal error and maximizes control confidence. For simplicity, we assume $P^X = p^X I_3$, where $p^X$ is a scalar constant and $I_3$ is the identity matrix of size 3. Using this assumption and taking the logarithmic determinant of Equation \ref{eqn:control_conf} after dropping the constant simplifies the control confidence term as:
\begin{equation} \label{eqn:logdet_Pi_u}
    \log |\Pi^U| =  \log \Big| \frac{\partial X}{\partial U}^\top \frac{\partial X}{\partial U}
+ (X - X^g)^\top  \frac{\partial^2 X}{\partial U^2} \Big|.
\end{equation}
From Equations \ref{eqn:free_energy} and \ref{eqn:logdet_Pi_u}, it can be seen that $p^X$ acts as a tuning parameter within $F$ to balance task performance and control confidence. Tool optimization in Equation \ref{eqn:opti_general_form} can be rewritten as:
\[
\mathcal{T}^* =  \underset{C \in \mathbb{R}^n}{\mathrm{argmin}} \quad \frac{1}{2} \sum_{t=0}^T
(X - X^g)^\top P^X (X - X^g) - \log |\Pi^U|, 
\]
\[
\text{subject to} \quad 
L(C) = L_0, \  
C = [c_1,\ldots, c_n, 0], \  
\pi_{\mathrm{plan}} = \pi_0,
\]
where $X$ is the trajectory of the box under control torques \(U\) of the robot - generated by executing the planned tool use policy \(\pi_{\mathrm{plan}}\). We use an evolutionary optimization strategy (CMAES) to generate a new population of tools and iteratively optimize it to minimize $F$. Depending on the value of $p^X$, the designed tool caters to the task completion or the control confidence.

\subsection{Tool design algorithm}
Our tool design algorithm has three main components: i) designer that optimizes and generates a tool, ii) user that executes a predetermined tool use in a robot simulation environment, and iii) an evaluator that judges the tool's performance in task completion and attributes a confidence for the tool. All three components iteratively interacts with each other towards an optimal tool design for the task. The schematic of our algorithm is given in Figure \ref{fig:schematic}, and the pseudocode is given in Alg. \ref{alg:design}. 

\begin{algorithm}
\caption{Autonomous Hand Tool Design via CMA-ES }
\begin{algorithmic}[1]
\State \textbf{Input:} Initial tool parameters $\mathcal{T}_0$, goal state $X^g$,
number of iterations $N$, plan $\pi_0$, CMA-ES parameters

\State \textbf{Initialize:} CMA-ES with mean $\mu = \mathcal{T}_0$, covariance $\Sigma$

\For{$k = 1$ to $N$}
    \State \textbf{Designer: Sample Tool Candidates}
    \For{$i = 1$ to population size}
        \State \quad Sample tool: $\mathcal{T}^{(i)} \sim \mathcal{N}(\mu, \Sigma)$
        \State \quad Tool mesh: $\text{mesh}^{(i)} \gets \text{make\_mesh}(\mathcal{T}^{(i)})$

        \State \textbf{User: Load Tool and Plan}
        \State \quad Load $\text{mesh}^{(i)}$ into simulator
        \State \quad Plan motion: $\pi^{(i)} \gets \pi_0$

        \State \textbf{User: Simulate with Controller}
        \State \quad $(X^{(i)}, U^{(i)}) \gets \text{simulate}( \pi^{(i)}, \mathcal{T}^{(i)})$

        \State \textbf{Evaluator: Compute Performance}
        \State \quad $E^{(i)} \gets \frac{1}{2} \sum_t (X^{(i)}_t - X^g)^\top P^X (X^{(i)}_t - X^g)$

        \State \textbf{Evaluator: Compute Confidence}
        \State \quad $\frac{\partial X}{\partial U}, \frac{\partial^2 X}{\partial U^2} \gets \text{numerical\_gradients} (X^{i},U^{i}) $
        \State \quad $C^{(i)} \gets - \frac{1}{2} \sum_t \log |\Pi^{(i)}_t| \gets \text{Equation } \ref{eqn:logdet_Pi_u}$

        \State \textbf{Evaluator: Compute Fitness}
        \State \quad $f^{(i)} \gets   E^{(i)} +   C^{(i)}$
    \EndFor

    \State \textbf{Designer: CMA-ES Update}
    \State \quad Update $\mu, \Sigma$ using $\{f^{(i)}\}$ and CMA-ES rules

\EndFor

\State \textbf{Return:} Optimized tool parameters $\mathcal{T}^*$

\end{algorithmic}  \label{alg:design}
\end{algorithm}

\section{RESULTS}

\subsection{Simulation setup}
The simulation is set up in Pybullet environment with a Franka Panda robot simulated to perform a sweeping and pulling motion, carrying the tool, aimed at bringing a box kept on the floor to a goal location, as shown in Figure \ref{fig:teaser}b.

\subsubsection{Tool parametrization and generation}
We take the minimalist parameterization necessary to show the effectiveness of our method: a stick-like tool with fixed curve length that is parameterized by a third-order polynomial of the form $c_1 x^3 + c_2x^2+c_3x+c_4$. This polynomial extrudes out of a square until a fixed curve length is achieved to form the full parametric mesh model. A straight segment is added at the end of the tool to standardize the robot tool grasping. To ensure continuity between the straight and curved segments, $c_4$ is constrained to 0. The tool generator generates the parametric model of the tool, given the polynomial coefficients $C=[c_1, c_2, c_3, 0]$. The examples of generated tools are given in Figure \ref{fig:teaser}.


\subsubsection{Motion planning and control}
Taking motivation from the idea that tool invention involves a known tool use, we fix the motion plan of the robot arm throughout the entire tool design process. The robot uses the following plan $\pi_0$: i) grasp and lift the tool to position $(q_a,0,h)$, ii) sweep motion: move the end effector to position $(0,q_a,h)$ in a straight line, and iii) pull motion: move the end effector to position $(0,q_b,h)$ in a straight line. This motion is expected to move a cuboid box of dimension $.1m\times.1m\times.2m$ and mass $m=.1kg$ to the goal location $X^g=[0, 1.5, .1]$. The tool is kept horizontal throughout. Although the controller and robot motion plan remain the same for the entire tool design routine, the control signals (robot joint torques) will differ depending on the moment of inertia of the tool.

\subsection{CMAES optimization}
A sample tool optimization process is shown in Figure \ref{fig:cmaes_progress}. The free energy (with $p^X = 40$) in Equation \ref{eqn:free_energy} is minimized after 15 iterations. The intermediate best polynomial tool curves and their performance in moving the box to the goal are shown in Figure \ref{fig:cmaes_progress} a and c, respectively. The end point of the box trajectories (black cross) in Figure \ref{fig:cmaes_progress}c can be observed to be close to the goal (red star), indicating a successful task. The resulting tool design (Figure \ref{fig:cmaes_progress}d) is mostly straight, such that the tool pushes the box away to the otherwise unreachable goal (under the given plan). This shows the success of our algorithm in optimizing the design of a tool using CMAES (with free energy as the objective) for task completion.

\begin{figure}[!hbtp] 
\centering
\includegraphics[scale = 0.265]{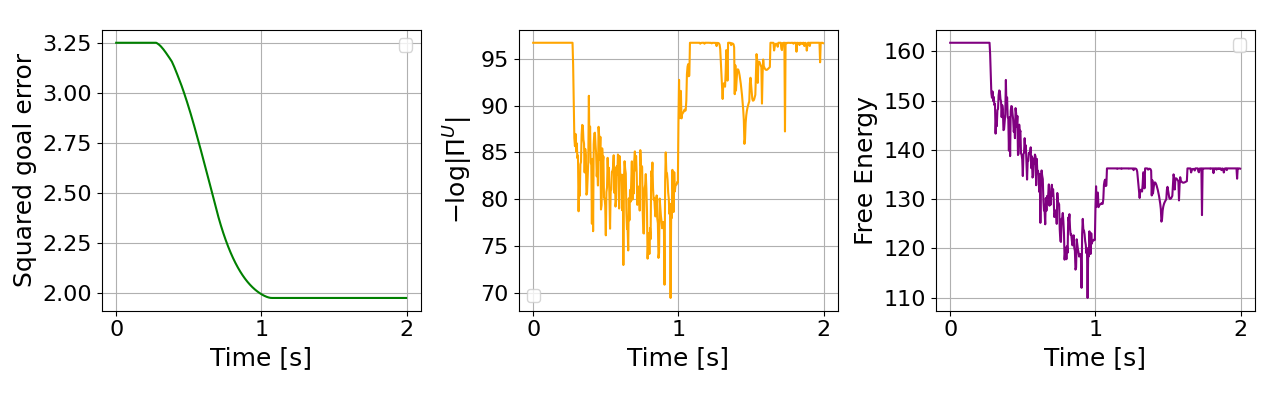}
\caption{The free energy objective rewards a decrease in goal error (before $t=1s$) and penalizes the decreases in tool's controllability on the box (after $t=1s$).}
\label{fig:FE_err_conf_time}
\end{figure}

\subsection{Control confidence for tool controllability}
This section highlights the importance of two contributing terms within $F$ in Equation \ref{eqn:free_energy}: i) squared goal error ($(X-X^g)^T(X-X^g)$), and ii) negative (scalar) control confidence $-\log|\Pi^U|$. They are plotted for a sample tool use in Figure \ref{fig:FE_err_conf_time}. The green curve is constant after $t=1s$, indicating that the box is stationary, and that the tool has lost control of the box. However, the (negative) control confidence term (in yellow) takes a higher value after $t=1s$, indicating that it is penalized for losing the tool's controllability on the box. This is in line with the role of control confidence, explained in Section \ref{sec:control_conf}. The noise in the graph is due to the discontinuous joint torques entering the numerical gradient computations in Equation \ref{eqn:logdet_Pi_u}. The free energy curve (in violet) can be seen as a weighted balance between squared goal error and confidence. This shows the effectiveness of our algorithm in using $F$ to reward task performance and penalize the loss in controllability of the tool for the task.

\begin{figure}[!hbtp]
\centering
\includegraphics[scale = 0.3]{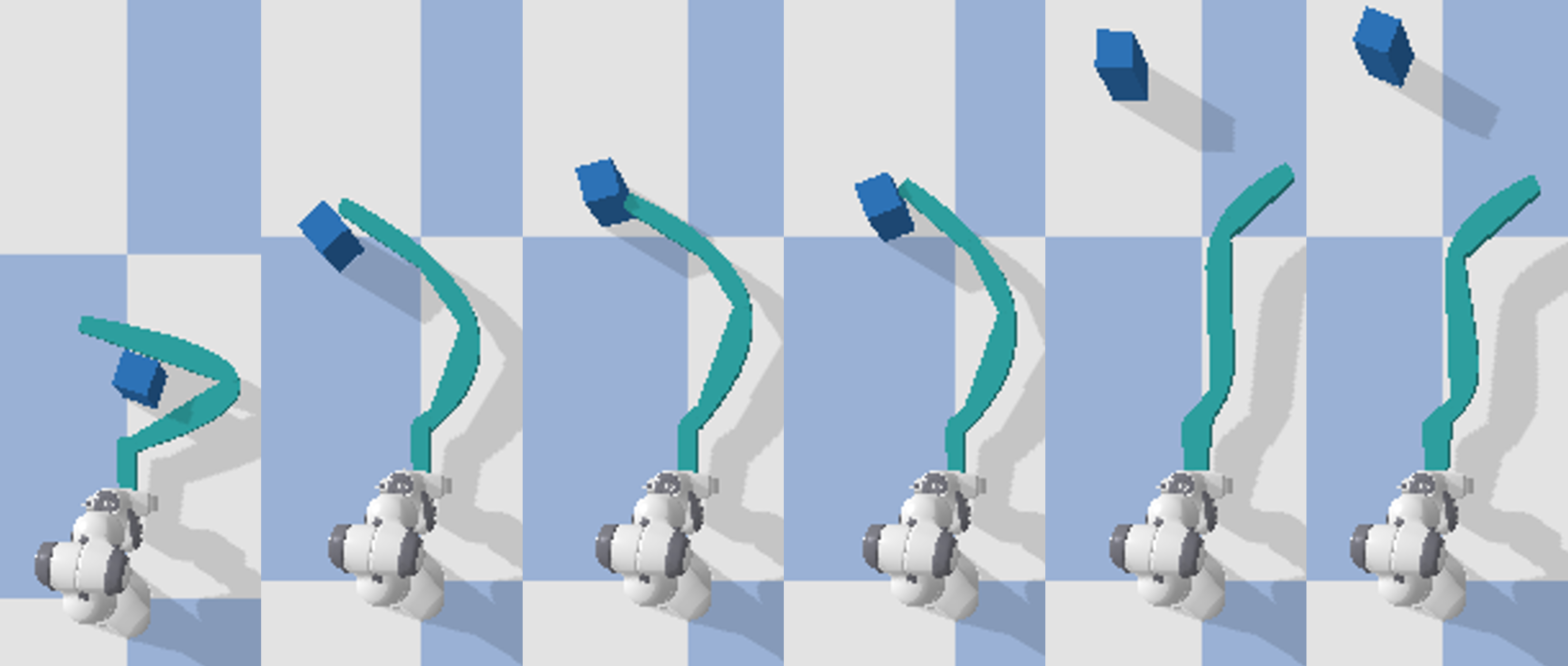}
\caption{The tools designed by the objectives with $p^X = 0,10,20,30,40,50$. The tool designed exclusively by optimizing control confidence ($p^X=0$) is highly curved and perfectly encloses the box for higher controllability, while the tool designed by squared error ($p^X=50$) is straighter and prioritize to push the box closer to the goal at the expense of controllability. }
\label{fig:tool_design_variations}
\end{figure}

\subsection{Tool design for performance and control confidence}
Figure \ref{fig:tool_design_variations} shows the tool designs that the free energy objectives with different weights ($p^X$) come up with. The tool designed for confidence in control (low $p^X$) is highly curved to enclose the box for better controllability while in use. However, tools designed for task performance (high $p^X$) are more straight and aim to push the box closer to the goal at the expense of controllability. The intermediate weights balance performance and confidence. Therefore, our algorithm can successfully design tools for high task performance and high confidence, depending on the values of $p^X$.

\begin{figure}[!hbtp]
\centering
\includegraphics[scale = 0.5]{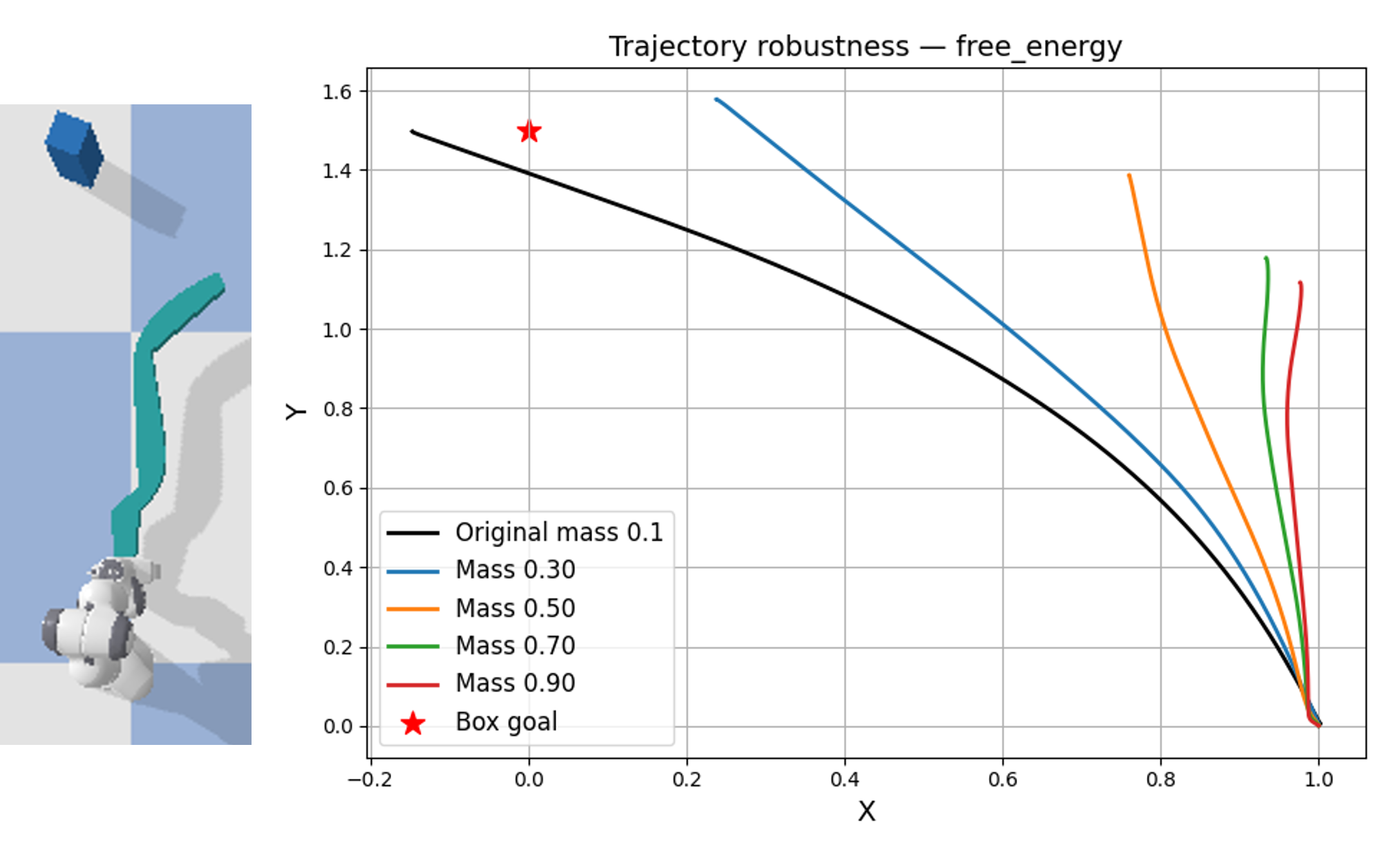}
\caption{Trajectory analysis of a tool with low control confidence weight $p^U=50$ (displayed on the left) when subjected to box mass perturbations. The box mass significantly affects the trajectory, departing from that of the box trajectory of the original box mass (black), with the maximum deviation shown by the highest mass (red).}
\label{fig:mass_pert_error_tool}
\end{figure}

\begin{figure}[!hbtp]
\centering
\includegraphics[scale = 0.5]{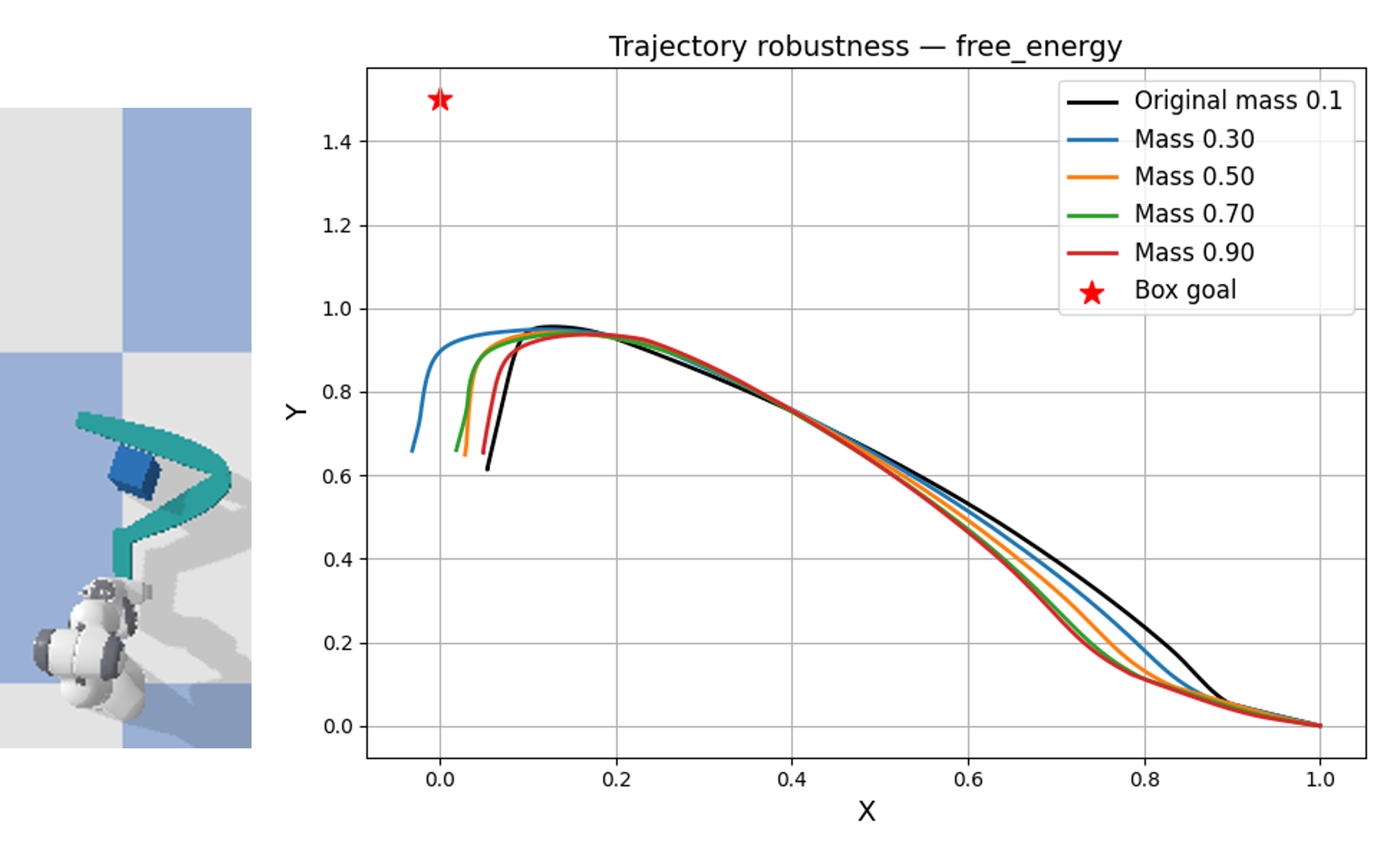}
\caption{ The tool designed by high control confidence weighting ($p^U = 0$ shown on the left side) shows high robustness to box mass perturbations. All curves stay close to the trajectory of the original mass (in black). }
\label{fig:mass_pert_conf_tool}
\end{figure}

\subsection{Uncertainty handling capability of tools}
This section aims to evaluate the uncertainty handling capability of tools designed using pure performance and pure confidence driven objectives. We model the mass uncertainty of the box as a simple case of environment uncertainty to test the tool's capabilities. The designed tools are tested under the same conditions, except for a change in box masses. Instead of the original mass $m=0.1kg$, the mass is altered ($m_i=\{0.3,0.5, 0.7, 0.9\}$), and the corresponding box trajectories are plotted in Figure \ref{fig:mass_pert_error_tool} and \ref{fig:mass_pert_conf_tool} for tools designed with pure performance ($p^X=50$) and pure confidence ($p^X=0$) respectively. It can be seen that the tool designed with pure confidence objective shows the least deviation in its box trajectories (curves in Figure \ref{fig:mass_pert_conf_tool} stay close to the black curve), compared to the tool designed with pure performance (curves in Figure \ref{fig:mass_pert_error_tool} deviate from the black curve). This shows that control confidence tools are more robust to environmental uncertainties than pure performance tools. The following sections will establish this fact using randomized simulation experiments.

\begin{figure}[!hbtp]
\centering
\includegraphics[scale = 0.4]{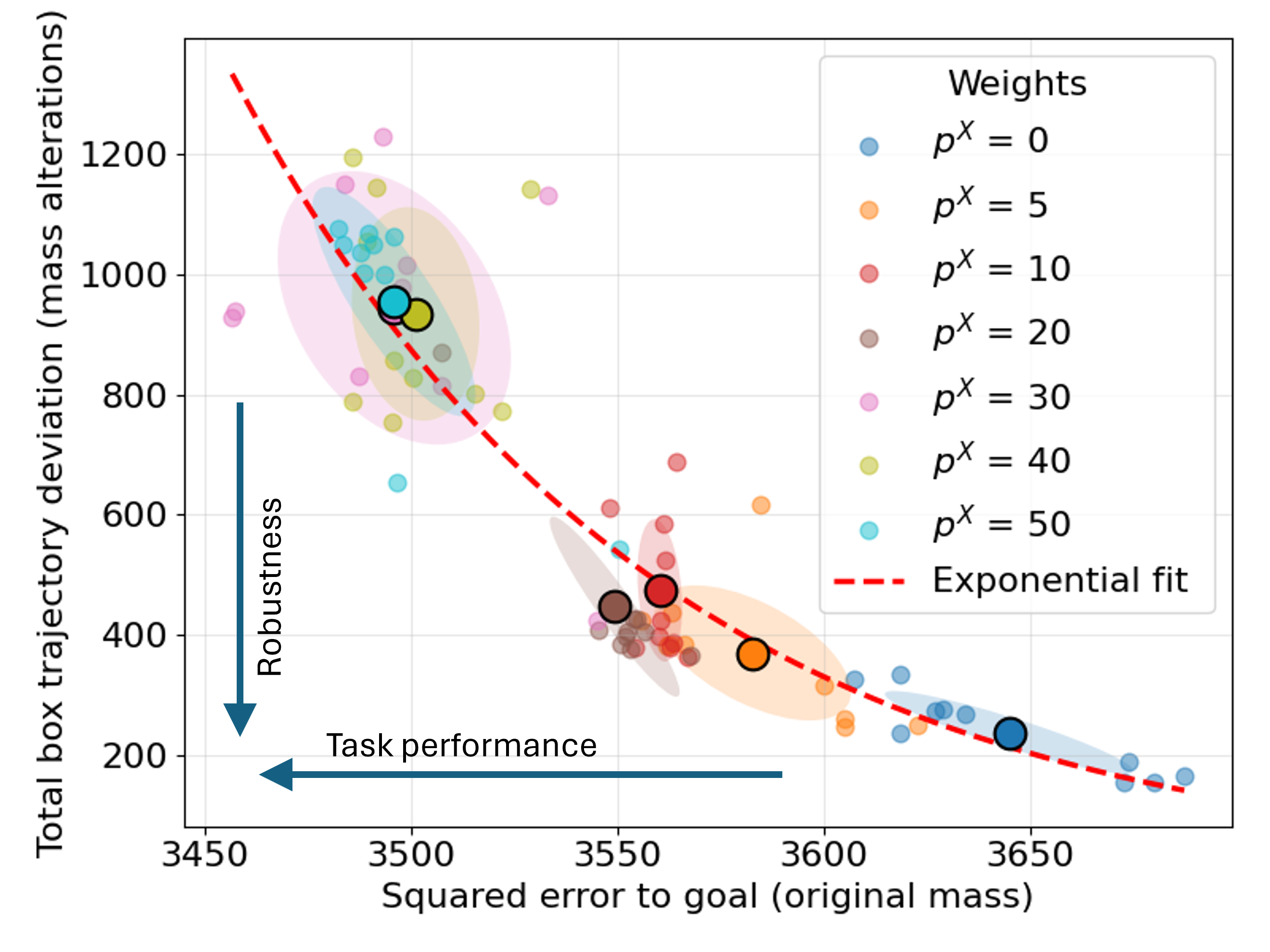}
\caption{The robustness of the tools designed by free energy objectives with different $p^X$. The tool designed by control confidence objective (weight = 0) shows the highest robustness (least box trajectory deviations), while the tool designed by squared error objective (weight = 50) shows the lowest robustness.  }
\label{fig:robustness_last70}
\end{figure}

\subsection{Robustness tests for designed tools }
To evaluate the robustness of the tool against environmental uncertainties, we define a metric that computes the deviation of the object trajectory when we change an environmental parameter (i.e., the object mass) from the one that has been optimized. Thus, it measures how reliable the tool design is to maintain the trajectory. For instance, in the push and pull task, a tool with higher robustness to mass perturbations will have a lower deviation in the box trajectory when the box mass is altered. Mathematically, our robustness measure is given by,
\begin{equation}
    robustness = - \sum_{i=1}^k \sum_{t=0}^T  || X(m_i,t) - X(m,t) ||,
\end{equation}
where $X(m_i,t)$ is the position of the box with perturbed (different) mass $m_i$ at time $t$, and $X(m,t)$ is the trajectory of the box with the original mass $m$. Figure \ref{fig:robustness_last70} shows the robustness of the tool (under mass alterations) plotted against the squared error to goal---trajectory with the original mass used for optimization---for 10 tool designs each for 7 different $p^X$ values. The results show that tools designed for confidence (low $p^X$) show high robustness to mass uncertainties at the expense of task performance (higher goal error), whereas tools designed for task performance (high $p^X$) show high task performance (low goal error) at the expense of robustness (high box trajectory deviations). Therefore, $p^X$ can be used as a tuning parameter within $F$ to balance the robustness of the tool (under mass alterations) and its task performance.

\subsection{Validating the evaluator}
We statistically evaluate the robustness-accuracy trade-off within the proposed free energy objective when optimizing the tool design. Besides robustness, we also define two more metrics: i) accuracy as the tool's task performance under object mass uncertainties, and ii) control deviation, which is analogous to robustness but for the control torques.

The tool that brings the box as quickly as possible to the goal position (even under mass alterations) will have a higher goal accuracy measure than a tool that takes the box away from the goal position. Mathematically, it is computed as the negative sum of the squared error of the box position with respect to the goal position for the whole trajectory, under all mass alterations, given by:
\begin{equation}
    accuracy = -\sum_{i=1}^k \sum_{t=0}^T || X(m_i,t) - X^g ||
\end{equation}
Similarly, we define the control deviation as follows:
\begin{equation}
    control \ deviation =  \sum_{i=1}^k \sum_{t=0}^T  || U(m_i,t) - U(m,t) ||,
\end{equation}
where $U(m_i,t) $ is the joint torque during tool use with the altered box mass $m_i$, and $U(m,t)$ is that with the original mass $m$.

\begin{table}[h!] 
\centering
\caption{Comparative analysis of different objectives on the robustness, accuracy and control deviation of the designed tool under box mass changes. } \label{tab:robustness_summary}
\begin{tabular}{c c c c c} 
\toprule
Objective & robustness $\uparrow$ & accuracy $\uparrow$ & control deviation $\downarrow$ \\
\midrule
Pure confidence & \textbf{-2008} & -65690& \textbf{58089} \\
Free energy & -2495 & \textbf{-64988} &  58116\\
Pure performance & -3384 & {-65130} & {59590} \\
\bottomrule
\end{tabular}
\end{table}

We perform a comparative analysis on a wide range of tool designs under three objective variants: i) pure confidence designs ($p^X=0$), ii) free energy designs ($p^X=20$), and iii) pure performance designs ($p^X=50$). By changing the goal location and fixing the motion plan, we obtain different optimized tool designs for each goal. We sample 10 random goal locations $X^g = [x^g, y^g, z^g]^T$ within the range $x^g=[-0.5,0.5],\ y^g=[0.5,1.5],\ z^g=0.1$ and use it for tool design for all three objectives for a fair comparison. 
Table \ref{tab:robustness_summary} shows the final robustness, goal accuracy and control deviations under mass alterations ($m_i=\{0.3,0.5, 0.7, 0.9\}$). The pure confidence objective generates tools with the highest robustness and the lowest control deviation under uncertainty (box mass), at the expense of accuracy. The free energy objective generates tools with the highest goal accuracy with better robustness and control deviation than the pure performance objective. The fragility of pure performance-based tools towards environmental changes can be countered by using $F$ as the objective to i) increase the tool's robustness and accuracy and ii) decrease the control deviation.  Therefore, the free-energy objective provides a sweet spot for robustness and accuracy for tool design.

\subsection{Validating the designer optimizer}
\begin{figure}[!hbtp]
\centering
\includegraphics[scale = 0.45]{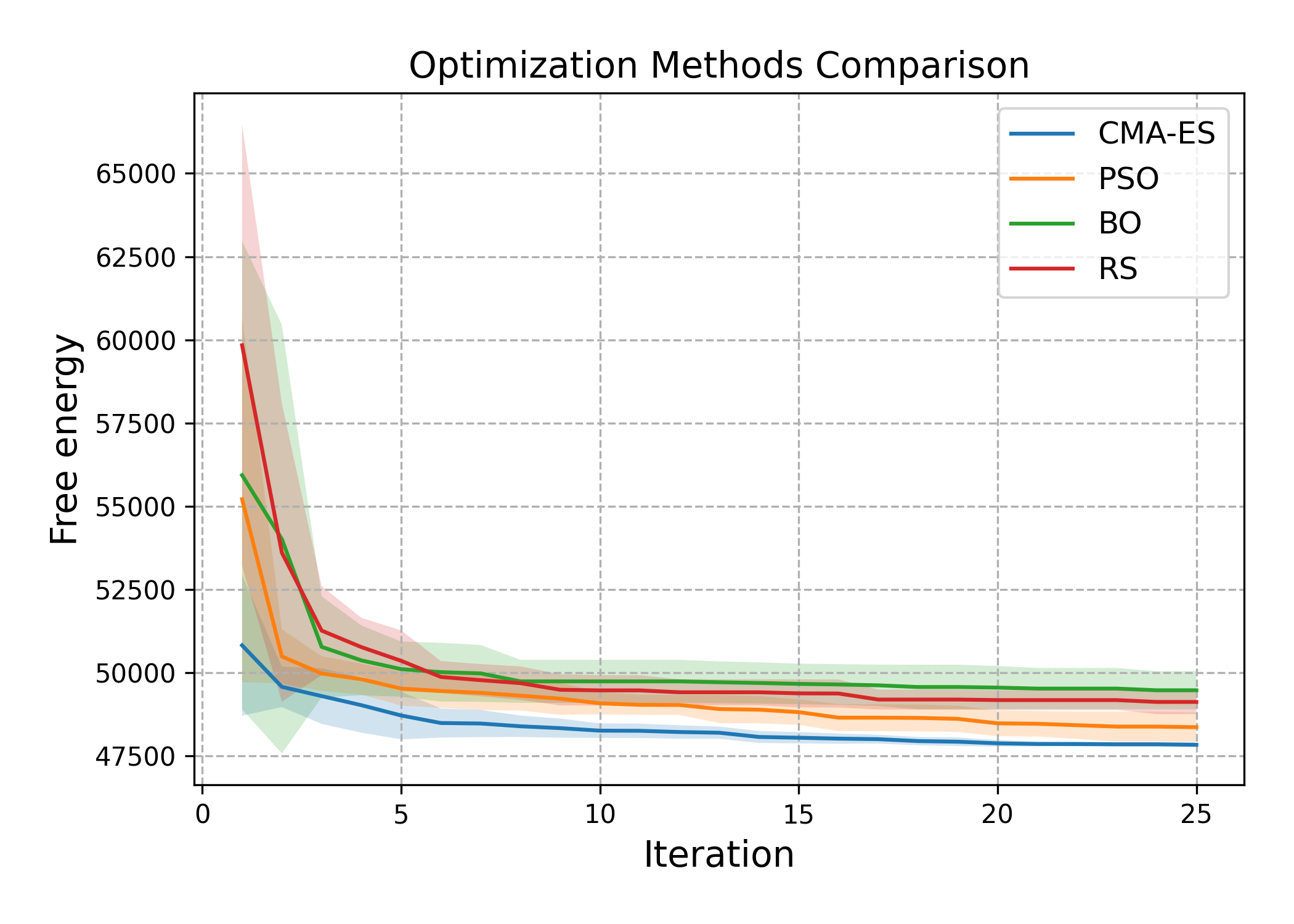}
\caption{The average decrease in free energy objective (with shaded uncertainty bounds) over iterations of the tool design using different optimizers.}
\label{fig:opt_comparison_FE_weight}
\end{figure}

The free energy landscape for our tool design problem is nonconvex and noisy due to the imperfections in mesh generation and the existence of multiple local optimal tool designs. Therefore, we benchmark the optimizer against other state-of-the-art optimizers used in robot simulations like Bayesian Optimization (BO), Particle Swarm Optimization (PSO) and Random Sampling (RS). For a fair comparison, all benchmarks are run under similar settings (population size, iterations, etc.) for 10 trials each. The results plotted in Figure \ref{fig:opt_comparison_FE_weight} show that CMAES (blue curve) drops the objective function (free energy objective with $p^x = 20$) the fastest on average compared to other optimizers (PSO, BO, random sampling), strengthening our optimizer choice for Alg. \ref{alg:design}. 

\section{CONCLUSIONS}
In this paper, we introduced the idea of using control confidence for hand tool design to improve the tool's robustness under environmental alterations. We described a general framework for tool design that includes confidence (i.e., second-order decision making), incorporating a key skill that enables humans to evaluate their confidence in performing a particular task (i.e., metacognitive performance). Hence, we introduced a mathematical account to implement metacognitive robots~\cite{della2024awareness} from a control perspective. Through rigorous simulation experiments, we showed that tools designed for control confidence are more robust to changes in the mass of the object in a reach, pushing, and pulling task. We further showed that using free energy as the objective for tool design seamlessly balances the goal accuracy and tool robustness. In terms of design complexity, future research can forgo the assumption of known tool use (fixed robot motion plan). Future work can focus on adaptively learning the tool use alongside tool design for complicated tasks like throwing, hammering, sawing, etc. Confidence-based decision making can be incorporated into other robotic applications, such as estimation, control, and discrete decision making. 

\section*{ACKNOWLEDGMENT}
Figure \ref{fig:teaser}a was created with the help of Adobe Firefly.






\bibliographystyle{IEEEtran}
\footnotesize
\bibliography{main}

\begin{thebibliography}{10}
\providecommand{\url}[1]{#1}
\csname url@samestyle\endcsname
\providecommand{\newblock}{\relax}
\providecommand{\bibinfo}[2]{#2}
\providecommand{\BIBentrySTDinterwordspacing}{\spaceskip=0pt\relax}
\providecommand{\BIBentryALTinterwordstretchfactor}{4}
\providecommand{\BIBentryALTinterwordspacing}{\spaceskip=\fontdimen2\font plus
\BIBentryALTinterwordstretchfactor\fontdimen3\font minus \fontdimen4\font\relax}
\providecommand{\BIBforeignlanguage}[2]{{%
\expandafter\ifx\csname l@#1\endcsname\relax
\typeout{** WARNING: IEEEtran.bst: No hyphenation pattern has been}%
\typeout{** loaded for the language `#1'. Using the pattern for}%
\typeout{** the default language instead.}%
\else
\language=\csname l@#1\endcsname
\fi
#2}}
\providecommand{\BIBdecl}{\relax}
\BIBdecl

\bibitem{qin2023robot}
M.~Qin, J.~Brawer, and B.~Scassellati, ``Robot tool use: A survey,'' \emph{Frontiers in Robotics and AI}, vol.~9, p. 1009488, 2023.

\bibitem{li_learning_2023}
\BIBentryALTinterwordspacing
M.~Li, R.~Antonova, D.~Sadigh, and J.~Bohg, ``Learning tool morphology for contact-rich manipulation tasks with differentiable simulation,'' in \emph{2023 IEEE International Conference on Robotics and Automation (ICRA)}.\hskip 1em plus 0.5em minus 0.4em\relax IEEE, May 2023, p. 1859–1865. [Online]. Available: \url{http://dx.doi.org/10.1109/icra48891.2023.10161453}
\BIBentrySTDinterwordspacing

\bibitem{liu_learning_2023}
\BIBentryALTinterwordspacing
Z.~Liu, S.~Tian, M.~Guo, C.~K. Liu, and J.~Wu, ``Learning to design and use tools for robotic manipulation,'' 2023. [Online]. Available: \url{https://arxiv.org/abs/2311.00754}
\BIBentrySTDinterwordspacing

\bibitem{yuan_transform2act_2022}
Y.~Yuan, Y.~Song, Z.~Luo, W.~Sun, and K.~Kitani, ``Transform2act: Learning a transform-and-control policy for efficient agent design,'' \emph{arXiv preprint arXiv:2110.03659}, 2021.

\bibitem{amant2005tool}
R.~S. Amant and A.~B. Wood, ``Tool use for autonomous agents.'' in \emph{AAAI}, 2005, pp. 184--189.

\bibitem{fleming2024metacognition}
S.~M. Fleming, ``Metacognition and confidence: A review and synthesis,'' \emph{Annual Review of Psychology}, vol.~75, no.~1, pp. 241--268, 2024.

\bibitem{meera2024confidence}
A.~A. Meera and P.~Lanillos, ``Confidence-aware decision-making and control for tool selection,'' in \emph{2024 IEEE/RSJ International Conference on Intelligent Robots and Systems (IROS)}.\hskip 1em plus 0.5em minus 0.4em\relax IEEE, 2024, pp. 12\,617--12\,624.

\bibitem{friston2010free}
K.~Friston, ``The free-energy principle: a unified brain theory?'' \emph{Nature reviews neuroscience}, vol.~11, no.~2, pp. 127--138, 2010.

\bibitem{lanillos2021active}
P.~Lanillos, C.~Meo, C.~Pezzato, A.~A. Meera, M.~Baioumy, W.~Ohata, A.~Tschantz, B.~Millidge, M.~Wisse, C.~L. Buckley \emph{et~al.}, ``Active inference in robotics and artificial agents: Survey and challenges,'' \emph{arXiv preprint arXiv:2112.01871}, 2021.

\bibitem{wu2021deepcaddeepgenerativenetwork}
\BIBentryALTinterwordspacing
R.~Wu, C.~Xiao, and C.~Zheng, ``Deepcad: A deep generative network for computer-aided design models,'' 2021. [Online]. Available: \url{https://arxiv.org/abs/2105.09492}
\BIBentrySTDinterwordspacing

\bibitem{wicaksono_towards_2017}
\BIBentryALTinterwordspacing
H.~Wicaksono, ``Towards a relational approach for tool creation by robots,'' in \emph{Proceedings of the Twenty-Sixth International Joint Conference on Artificial Intelligence}, ser. IJCAI-2017.\hskip 1em plus 0.5em minus 0.4em\relax International Joint Conferences on Artificial Intelligence Organization, Aug. 2017, p. 5221–5222. [Online]. Available: \url{http://dx.doi.org/10.24963/ijcai.2017/770}
\BIBentrySTDinterwordspacing

\bibitem{wicaksono_cognitive_2020}
\BIBentryALTinterwordspacing
H.~Wicaksono and C.~Sammut, ``A cognitive robot equipped with autonomous tool innovation expertise,'' \emph{International Journal of Electrical and Computer Engineering (IJECE)}, vol.~10, no.~2, p. 2200, Apr. 2020. [Online]. Available: \url{http://dx.doi.org/10.11591/ijece.v10i2.pp2200-2207}
\BIBentrySTDinterwordspacing

\bibitem{yang_autonomous_2020}
\BIBentryALTinterwordspacing
C.~Yang, X.~Lan, H.~Zhang, and N.~Zheng, ``Autonomous tool construction with gated graph neural network,'' in \emph{2020 IEEE International Conference on Robotics and Automation (ICRA)}.\hskip 1em plus 0.5em minus 0.4em\relax IEEE, May 2020, p. 9708–9714. [Online]. Available: \url{http://dx.doi.org/10.1109/ICRA40945.2020.9197285}
\BIBentrySTDinterwordspacing

\bibitem{choi_creating_nodate}
D.~Choi, P.~Langley, and S.~T. To, ``Creating and using tools in a hybrid cognitive architecture.'' in \emph{AAAI Spring Symposia}, 2018.

\bibitem{nair_tool_2019}
\BIBentryALTinterwordspacing
L.~Nair, J.~Balloch, and S.~Chernova, ``Tool macgyvering: Tool construction using geometric reasoning,'' in \emph{2019 International Conference on Robotics and Automation (ICRA)}.\hskip 1em plus 0.5em minus 0.4em\relax IEEE, May 2019, p. 5837–5843. [Online]. Available: \url{http://dx.doi.org/10.1109/ICRA.2019.8793257}
\BIBentrySTDinterwordspacing

\bibitem{nair_feature_2020}
\BIBentryALTinterwordspacing
L.~Nair and S.~Chernova, ``Feature guided search for creative problem solving through tool construction,'' \emph{Frontiers in Robotics and AI}, vol.~7, Dec. 2020. [Online]. Available: \url{http://dx.doi.org/10.3389/frobt.2020.592382}
\BIBentrySTDinterwordspacing

\bibitem{wu_imagine_2020}
Y.~Wu, S.~Kasewa, O.~Groth, S.~Salter, L.~Sun, O.~P. Jones, and I.~Posner, ``Imagine that! leveraging emergent affordances for 3d tool synthesis,'' \emph{arXiv preprint arXiv:1909.13561}, 2019.

\bibitem{li_generating_2024}
\BIBentryALTinterwordspacing
M.~Li, L.~Kong, and S.~Kriegman, ``Generating freeform endoskeletal robots,'' 2024. [Online]. Available: \url{https://arxiv.org/abs/2412.01036}
\BIBentrySTDinterwordspacing

\bibitem{lin_robotsmith_2025}
\BIBentryALTinterwordspacing
C.~Lin, H.~Yuan, Y.~Wang, X.~Qiu, T.-H. Wang, M.~Guo, B.~Wang, Y.~Narang, D.~Fox, and C.~Gan, ``Robotsmith: Generative robotic tool design for acquisition of complex manipulation skills,'' 2025. [Online]. Available: \url{https://arxiv.org/abs/2506.14763}
\BIBentrySTDinterwordspacing

\bibitem{gao_vlmgineer_2025}
\BIBentryALTinterwordspacing
G.~J. Gao, T.~Li, J.~Shi, Y.~Li, Z.~Zhang, N.~Figueroa, and D.~Jayaraman, ``Vlmgineer: Vision language models as robotic toolsmiths,'' 2025. [Online]. Available: \url{https://arxiv.org/abs/2507.12644}
\BIBentrySTDinterwordspacing

\bibitem{pun_generating_2025}
\BIBentryALTinterwordspacing
A.~Pun, K.~Deng, R.~Liu, D.~Ramanan, C.~Liu, and J.-Y. Zhu, ``Generating physically stable and buildable brick structures from text,'' 2025. [Online]. Available: \url{https://arxiv.org/abs/2505.05469}
\BIBentrySTDinterwordspacing

\bibitem{seita_toolflownet_2023}
D.~Seita, Y.~Wang, S.~J. Shetty, E.~Y. Li, Z.~Erickson, and D.~Held, ``Toolflownet: Robotic manipulation with tools via predicting tool flow from point clouds,'' in \emph{Conference on Robot Learning}.\hskip 1em plus 0.5em minus 0.4em\relax PMLR, 2023, pp. 1038--1049.

\bibitem{collis_understanding_2024}
\BIBentryALTinterwordspacing
P.~Collis, P.~F. Kinghorn, and C.~L. Buckley, \emph{Understanding Tool Discovery and Tool Innovation Using Active Inference}.\hskip 1em plus 0.5em minus 0.4em\relax Springer Nature Switzerland, Nov. 2023, p. 43–58. [Online]. Available: \url{http://dx.doi.org/10.1007/978-3-031-47958-8_4}
\BIBentrySTDinterwordspacing

\bibitem{xu_dynamics-guided_2025}
\BIBentryALTinterwordspacing
X.~Xu, H.~Ha, and S.~Song, ``Dynamics-guided diffusion model for sensor-less robot manipulator design,'' 2024. [Online]. Available: \url{https://arxiv.org/abs/2402.15038}
\BIBentrySTDinterwordspacing

\bibitem{lu_dynamic_2025}
\BIBentryALTinterwordspacing
Z.~Lu, N.~Wang, and C.~Yang, ``A dynamic movement primitives-based tool use skill learning and transfer framework for robot manipulation,'' \emph{IEEE Transactions on Automation Science and Engineering}, vol.~22, p. 1748–1763, 2025. [Online]. Available: \url{http://dx.doi.org/10.1109/TASE.2024.3370139}
\BIBentrySTDinterwordspacing

\bibitem{qi_learning_2024}
\BIBentryALTinterwordspacing
C.~Qi, Y.~Wu, L.~Yu, H.~Liu, B.~Jiang, X.~Lin, and D.~Held, ``Learning generalizable tool-use skills through trajectory generation,'' in \emph{2024 IEEE/RSJ International Conference on Intelligent Robots and Systems (IROS)}.\hskip 1em plus 0.5em minus 0.4em\relax IEEE, Oct. 2024, p. 2847–2854. [Online]. Available: \url{http://dx.doi.org/10.1109/iros58592.2024.10801653}
\BIBentrySTDinterwordspacing

\bibitem{li_roman_2022}
\BIBentryALTinterwordspacing
J.~Li, A.~Samoylov, J.~Kim, and X.~A. Chen, ``Roman: Making everyday objects robotically manipulable with 3d-printable add-on mechanisms,'' in \emph{CHI Conference on Human Factors in Computing Systems}, ser. CHI ’22.\hskip 1em plus 0.5em minus 0.4em\relax ACM, Apr. 2022, p. 1–17. [Online]. Available: \url{http://dx.doi.org/10.1145/3491102.3501818}
\BIBentrySTDinterwordspacing

\bibitem{kodnongbua_computational_2022}
\BIBentryALTinterwordspacing
M.~Kodnongbua, I.~Good, Y.~Lou, J.~Lipton, and A.~Schulz, ``Computational design of passive grippers,'' \emph{ACM Transactions on Graphics}, vol.~41, no.~4, p. 2–12, Jul. 2022. [Online]. Available: \url{http://dx.doi.org/10.1145/3528223.3530162}
\BIBentrySTDinterwordspacing

\bibitem{fleming2017self}
S.~M. Fleming and N.~D. Daw, ``Self-evaluation of decision-making: A general bayesian framework for metacognitive computation.'' \emph{Psychological review}, vol. 124, no.~1, p.~91, 2017.

\bibitem{friston2008variational}
K.~J. Friston, N.~Trujillo-Barreto, and J.~Daunizeau, ``Dem: a variational treatment of dynamic systems,'' \emph{Neuroimage}, vol.~41, no.~3, pp. 849--885, 2008.

\bibitem{della2024awareness}
C.~Della~Santina, C.~H. Corbato, B.~Sisman, L.~A. Leiva, I.~Arapakis, M.~Vakalellis, J.~Vanderdonckt, L.~F. D’Haro, G.~Manzi, C.~Becchio \emph{et~al.}, ``Awareness in robotics: an early perspective from the viewpoint of the eic pathfinder challenge “awareness inside”,'' in \emph{European Robotics Forum}.\hskip 1em plus 0.5em minus 0.4em\relax Springer, 2024, pp. 108--113.

\end{thebibliography}

\end{document}